\documentclass[a4paper,conference]{IEEEtran}
\IEEEoverridecommandlockouts
\usepackage{amsmath,amssymb,amsfonts}
\usepackage[noend]{algorithmic}
\usepackage{graphicx}
\usepackage{mathtools}
\usepackage{textcomp}
\usepackage{pifont}
\usepackage{bbding}
\usepackage{xcolor}
\usepackage{algorithm}
\usepackage{booktabs,threeparttable}
\usepackage{adjustbox}
\usepackage{color,colortbl}
\usepackage{float}
\usepackage{multirow}
\usepackage{array}
\usepackage{makecell}
\usepackage{lipsum,setspace}
\usepackage{tabu}
\usepackage{romannum}
\usepackage[noadjust]{cite}
\def\BibTeX{{\rm B\kern-.05em{\sc i\kern-.025em b}\kern-.08em
    T\kern-.1667em\lower.7ex\hbox{E}\kern-.125emX}}
\begin{document}

\title{Few-Shot Object Detection with Proposal \\Balance Refinement}

\author{\IEEEauthorblockN{Sueyeon Kim$^1$, Woo-Jeoung Nam$^2$, Seong-Whan Lee$^{1,2}$}
\IEEEauthorblockA{1. Department of Artificial Intelligence, Korea University\\
2. Department of Computer and Radio Communications Engineering, Korea University \\
Email: \{sueyeon\_kim, nwj0612, sw.lee\}@korea.ac.kr}\\
}

\maketitle
\begin{abstract}
Few-shot object detection has gained significant attention in recent years as it has the potential to greatly reduce the reliance on large amounts of manually annotated bounding boxes. While most existing few-shot object detection literature primarily focuses on bounding box classification by obtaining as discriminative feature embeddings as possible, we emphasize the necessity of handling the lack of intersection-over-union (IoU) variations induced by a biased distribution of novel samples. 
In this paper, we analyze the IoU imbalance that is caused by the relatively high number of low-quality region proposals, and reveal that it plays a critical role in improving few-shot learning capabilities. The well-known two stage fine-tuning technique causes insufficient quality and quantity of the novel positive samples, 
which hinders the effective object detection of unseen novel classes.
To alleviate this issue, we present a few-shot object detection model with proposal balance refinement, a simple yet effective approach in learning object proposals using an auxiliary sequential bounding box refinement process. This process enables the detector to be optimized on the various IoU scores through additional novel class samples.
To fully exploit our sequential stage architecture, we revise the fine-tuning strategy and expose the Region Proposal Network to the novel classes in order to provide increased learning opportunities for the region-of-interest (RoI) classifiers and regressors. Our extensive assessments on PASCAL VOC and COCO demonstrate that our framework substantially outperforms other existing few-shot object detection approaches.
\end{abstract}


\section{Introduction}
Recent significant progress in deep neural networks has led to remarkable success in computer vision tasks\cite{lee1999integrated, lee2003pattern, yang2007reconstruction} such as image classification\cite{he2016deep, kipf2016semi,zoph2018learning,dosovitskiy2020image} and object detection\cite{girshick2015fast,ren2015faster,liu2016ssd,redmon2016you,redmon2017yolo9000,redmon2018yolov3,bochkovskiy2020yolov4, carion2020end, cai2018cascade,vu2021scnet, chen2019hybrid}. However, these improvements are only relevant when a large amount of annotated data is available, which is more difficult in the case of object detection. It takes extra effort and cost to annotate data, as it requires not only identifying the categorical labels for all objects in the image but also providing accurate location information with bounding box coordinates. 
Moreover, the ability of a human to grasp novel notions at a few glances is still beyond the capabilities of current models. There has been thus significant interest in identifying unseen objects given very few training examples.

Following the previous philosophy of few-shot image classification\cite{koch2015siamese, li2019large,snell2017prototypical, vinyals2016matching, garcia2017few, gidaris2018dynamic, finn2017model, matsumi2021few}, existing few-shot object detection approaches mainly focus on achieving as  discriminative feature
\begin{figure}[H]
\begin{center}
\includegraphics[width=0.95\linewidth]{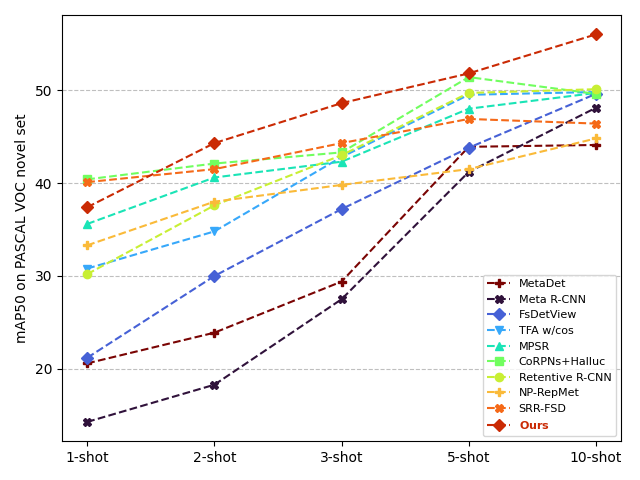}
\end{center}
\caption{Few-shot object detection performance (mAP) on the third PASCAL VOC\cite{everingham2010pascal} novel set at different shot numbers. The performance of our approach is superior to other state-of-the-art approaches.}
\end{figure}embeddings as possible to improve the bounding box classification performance. 
The recent state-of-the-art method SRR-FSD\cite{zhu2021semantic} further strengthens the discrimination among the classes with additional class-wise semantic embeddings. CoRPNs+Halluc\cite{zhang2021hallucination} expands the detector with a hallucinator to address the lack of variation in training data and boost the classification of novel classes. 
These prior works demonstrate that few-shot object detection can be alleviated by improving the classification performance of the Region-of-Interests (RoIs). Nevertheless, object detection is more challenging to be improved relying solely on attaining discriminative feature embeddings, since it is by definition a jointed task of classification and localization. The quality of region proposals is also a crucial factor to the overall object detection performance (the Intersection over Union (IoU) score). However, the current design of the fine-tuning based approach, TFA\cite{wang2020frustratingly}, induces a heavily biased IoU distribution of novel samples. In particular, TFA chooses to freeze a feature extractor and the region proposal network (RPN) during the novel fine-tuning stage, in order to avoid deteriorating a well-constructed feature space from the sufficient training samples. Necessarily, RPN has no choice but to directly facilitate the feature representation obtained from the base training phase to predict the objects of novel classes. However, this brings two major drawbacks: (1) The general quality of region proposals in novel fine-tuning stage is restricted to be lower than the base counterpart. (2) The number of proposals created from the positive anchors in novel fine-tuning stage is substantially less than the base counterpart. Our intuition is that this clear gap between the stages is one of the primary factors attributing to the current unsatisfactory performance of few-shot object detection.

To resolve this problem, we present few-shot object detection with proposal balance refinement approach, aiming at solving the imbalanced IoU distribution and deficient amount of good quality proposals of novel classes via exploiting auxiliary sequential refinement process. 
This simple resampling process provides us with a new solution to increase the number of proposals at different IoU degrees.
The contributions of this study are three-fold:
\begin{itemize}
    \item We carefully analyze and address the fundamental weakness of the fine-tuning based approach – optimizing the detector to the novel RoIs of the severely imbalanced IoU distribution which, to the best of our knowledge, has not been tackled before.
    \item To address this problem, we propose a new approach in few-shot domain to enrich the RoI samples to include various IoU scores, and balance out the disproportionate IoU distribution that are highly prevalent under the data-scarce scenarios.  
    \item In the verified experiential settings, we assess our model on the standard benchmark datasets: PASCAL VOC and COCO. The results demonstrate that our approach achieves a sizeable improvement compared to other existing methods without bells and whistles.
\end{itemize}

\section{Related Works}

\subsection{Object detection}
Object detection is one of the most fundamental problems in computer vision. There are two main lines in the current systems: two-stage proposal-based approach and one-stage integrated-manner one. 
R-CNN series\cite{ren2015faster, he2017mask} has the two-stage architecture of firstly predicting probable candidate regions of the image with multiple anchors per pixel in the RPN and then distinguishing the RoIs by category classification and bounding box localization. Amongst these proposal-based approaches, multi-stage detectors\cite{cai2018cascade,chen2019hybrid,vu2021scnet} show high performing detection and segmentation abilities by resampling the RoIs. 
Meanwhile, the one-stage object detectors including YOLO\cite{redmon2016you}, SSD\cite{liu2016ssd} and their variants\cite{redmon2017yolo9000, redmon2018yolov3, bochkovskiy2020yolov4} directly detect objects in the image with a single fully convolutional network. As all these frameworks are designed without the consideration of data-scarce scenarios, it is inappropriate to directly utilize them for recognizing unseen novel class objects.

\subsection{Few-shot Object Detection}
Recently, various meta-learning\cite{yan2019meta, wang2019meta, xiao2020few,li2021beyond} and metric learning\cite{yang2020restoring} based approaches have been proposed to tackle few-shot object detection. Meta R-CNN\cite{yan2019meta} reweights the 
\begin{figure}[H]
\begin{center}
\includegraphics[width=0.95\linewidth]{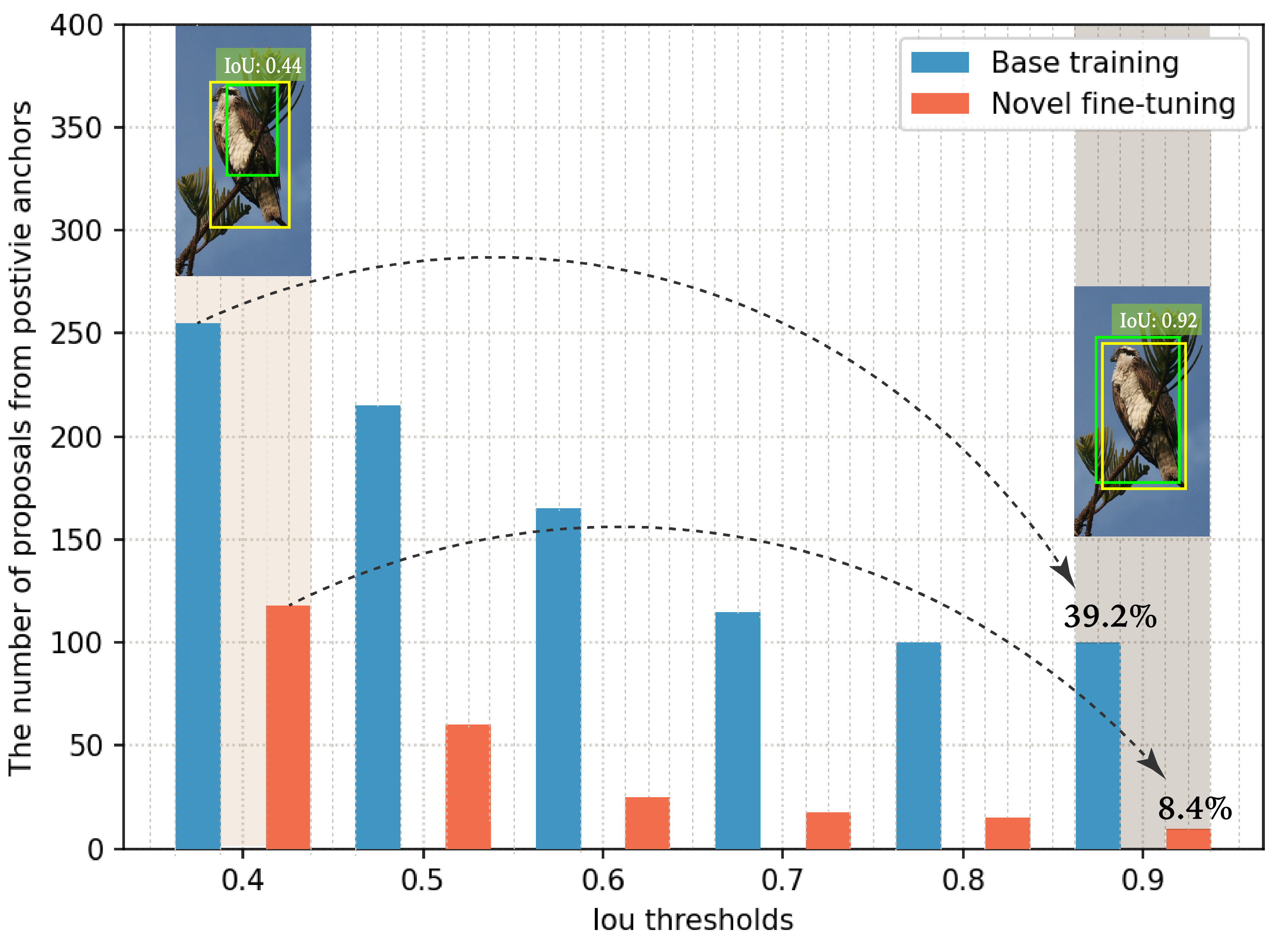}
\end{center} 
\caption{Imbalanced intersection-over-union (IoU) distribution under the few-shot setting. The blue and orange bars denote the average number of proposals from positive anchors of base training and novel fine-tuning, respectively. The yellow bounding boxes denote the ground-truth, while the green bounding boxes are the predicted results yielded by RPN. The percentages indicated above the bars of IoU threshold 0.9 represent the reduced ratio compared to the case of 0.4. As shown in the figure, there is a sizable gap between the two training steps. }
\end{figure}
importance of the attention of each RoI feature with the help of a meta-learner and utilizes the class attentive vectors. 
Meta-Det\cite{wang2019meta} focuses on predicting the parameters of category-specific components from few samples.
FsDetView\cite{xiao2020few} presents a joint feature embedding to leverage rich feature information from abundant base class examples and Np-RepMet\cite{yang2020restoring} proposes an inference scheme with negative and positive representatives by restoring negative information. CME\cite{li2021beyond} aims at retaining proper margin space among novel classes. 
However, these meta-learning based methods require specific data preparation, and have difficulty in achieving the conventional performance. Accordingly, fine-tuning based approaches\cite{wang2020frustratingly, zhang2021hallucination, fan2021generalized, wu2020multi} are increasingly becoming recognized in the field. TFA\cite{wang2020frustratingly} introduces a simple two-stage fine-tuning approach where the detector is firstly trained on all base classes with abundant samples and then fine-tuned only on the few samples of novel classes. Most of the fine-tuning based works follow the training scheme of TFA, improving upon it from various angles. MPSR\cite{wu2020multi} tackles the problem of scale variation and Retentive R-CNN\cite{fan2021generalized} debiases the pretrained RPN and emphasizes on the base forgetting issue. FSOD-SR\cite{kim2021spatial} and SRR-FSD\cite{zhu2021semantic} put forward the importance of exploiting the contextual information by considering the co-occurrence of objects in the visual scenes. CoRPNs+Halluc\cite{zhang2021hallucination} introduces a hallucinator to further augment the feature variation of few given samples.
As all these transfer-learning based approaches are the variants of TFA, our work also takes TFA as our main baseline and aims at resolving the remaining issue. We tackle the highly disproportionate IoU distribution of unseen novel classes via a sequential proposal refinement process.

\newcommand{\ie}{\textit{i}.\textit{e}\textit{.}\textit{,}}
\newcommand{\cmark}{\ding{51}}%
\newcommand{\xmark}{\ding{55}}%
\newcommand{\centered}[1]{\begin{tabular}{l} #1 \end{tabular}}
\makeatletter
\newcommand*{\da@rightarrow}{\mathchar"0\hexnumber@\symAMSa 4B }
\newcommand*{\da@leftarrow}{\mathchar"0\hexnumber@\symAMSa 4C }
\newcommand*{\xdashrightarrow}[2][]{%
  \mathrel{%
    \mathpalette{\da@xarrow{#1}{#2}{}\da@rightarrow{\,}{}}{}%
  }%
}
\newcommand*{\da@xarrow}[7]{%
  \sbox0{$\ifx#7\scriptstyle\scriptscriptstyle\else\scriptstyle\fi#5#1#6\m@th$}%
  \sbox2{$\ifx#7\scriptstyle\scriptscriptstyle\else\scriptstyle\fi#5#2#6\m@th$}%
  \sbox4{$#7\dabar@\m@th$}%
  \dimen@=\wd0 %
  \ifdim\wd2 >\dimen@
    \dimen@=\wd2 %
  \fi
  \count@=2 %
  \def\da@bars{\dabar@\dabar@}%
  \@whiledim\count@\wd4<\dimen@\do{%
    \advance\count@\@ne
    \expandafter\def\expandafter\da@bars\expandafter{%
      \da@bars
      \dabar@ 
    }%
  }%
  \mathrel{#3}%
  \mathrel{%
    \mathop{\da@bars}\limits
    \ifx\\#1\\%
    \else
      _{\copy0}%
    \fi
    \ifx\\#2\\%
    \else
      ^{\copy2}%
    \fi
  }%
  \mathrel{#4}%
}
\makeatother
 \section{Proposed method}
 We first introduce the problem settings of few-shot object detection in subsection A. Then we revisit the proposal imbalance issue in existing approaches in B and expand our proposal balance refinement approach in C.
 \subsection{Few-Shot Object Detection Setting}
In the standard setup of few-shot object detection from previous works [13, 21, 50, 55], classes are split into two sets: base classes $C_{base}$ and novel classes $C_{novel}$ where $C_{base}\cap\ C_{novel}=\O$. Accordingly, datasets $D=\{(x,y)|x\in X, y\in Y\}$ are composed of two types based on these classes: $S_{base}$ and $S_{novel}$ where $x$ is input image and $y=\{(c_i,b_i)|i=1,...,N\}$. Here, $c_i$ and $b_i$ denotes the category and the bounding box coordinates of the  $i$-th object out of the N object instances in the image $x$, respectively. The full training procedure consists of two-stage fine-tuning paradigm: the base training stage and novel fine-tuning stage. 
In the first base training stage, the model $Det$ is trained on a large base set $S_{base}$ and becomes a $|C_{base}|$-way detector.
In the novel fine-tuning stage, model is fine-tuned on a balanced set of $C_{base}\cup\ C_{novel}$, such that it can maintain the detection performance on pre-trained base classes and learn about novel classes at the same time. Accordingly, the model becomes a ($|C_{base}| + |C_{novel}|$)-way detector in the second stage and can be summarized as follow, 
\begin{equation}
    Det_{init} \xhookrightarrow{{S_{base}}} Det_{base}\xdashrightarrow{{S_{novel}}} Det_{novel}
\end{equation}
where $\hookrightarrow$ and $\dashrightarrow$ represents the conventional training and the fine-tuning process, respectively. The captions above the arrows denote the utilized dataset for training.

\subsection{Revisiting Proposal Imbalance}
The primary philosophy of TFA is to implicitly leverage the knowledge of multiple base classes to construct novel class feature space. Derived from the performance trade-off between novel classes and base classes, only the last RoI classifier and regressor are further fine-tuned on the data-scarce novel classes. Nevertheless, this gives rise to two negative consequences: inadequate quality and quantity of proposals from RPN during the novel-fine tuning.
One could assume RPN is a wholly class-agnositc component as it is merely trained to classify the objectness of the anchors. However, RPN inevitably becomes tilted towards the base classes because the anchors of novel class objects are categorized into non-object during the base training. 
Fig.2 shows that the number of proposals from the positive anchors decreases as the IoU threshold increases in both stages. However, the general quality of region proposals in novel fine-tuning stage is suppressed to be lower than the base counterpart. In particular, in the novel fine-tuning stage, the region proposals of IoU [0.4, 0.6) comprise of 72.4\%, whereas its base training counterpart comprises of only 49.4\%. 
Moreover, in the base training, the number of the highest quality proposals of IoU [0.9, 1.0) takes up 39.2\% of the counterpart of IoU [0.4, 0.5). However, in the novel-fine tuning stage, the highest-quality proposals are only 8.4\% of the lowest quality.
Besides, in terms of quantity, there is a large discrepancy between the two statistics in the whole IoU range. Setting the IoU threshold [0.5, 0.9], the amount of the proposals of novel fine-tuning is even less than 25\% of the base training. These two properties induce an undesirable consequence of interfering the improvement of learning capabilities on novel classes: Less and low-quality positive anchors become proposals, which is followed by the lack of learning opportunities for the RoI classifier and regressor. Necessarily, this inhibits the detector from obtaining various feature variations of novel classes and induces the overfitting to the proposals of impaired quality.
Based on these analyses above, we argue that these neglected issue of severe IoU imbalance and extremely deficient amount of the positive samples of novel classes should be resolved to boost the unsatisfactory detection performance of few-shot setting.
\begin{figure*}
\begin{center}
\small
\centerline{\includegraphics[width=0.9\textwidth]{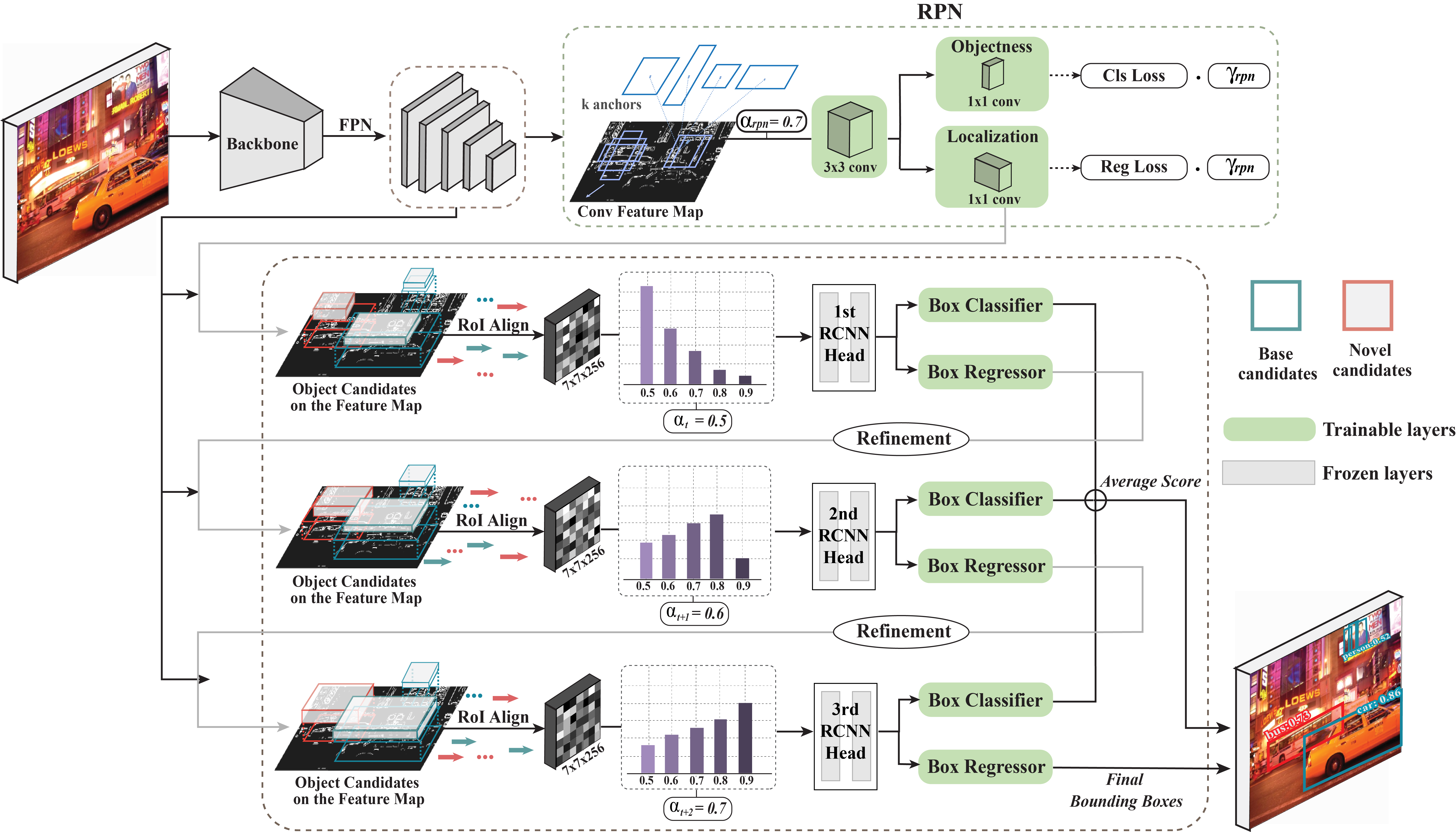}}
\end{center}
\caption{An overall architecture of our approach. Green and grey indicates the trainable layers and the frozen layers during the novel fine-tuning phase, respectively. $\alpha_{rpn}$ is the IoU threshold of RPN and $\alpha_t$ is the IoU threshold of each stage $t$ of the sequential detectors. $\alpha_t$ is set to increase as the stage ascends. The turquoise and orange bounding boxes represent the region proposals of novel classes and base classes. The grey arrow line indicates the flow of the regressed bounding boxes. The improvement of IoU distribution of RoIs at each stage is illustrated with bar graphs before the RCNN Heads.  $\gamma_{rpn}$ denotes the loss coefficient of the RPN.}

\end{figure*}

\subsection{Proposal Balance Refinement}
Motivated by the aforementioned findings, we employ an auxiliary proposal balance refinement process during the base training and novel fine-tuning. As shown in Fig. 3, a given image is firstly processed by the backbone network, FPN and RPN. Subsequently, it is fed to the sequential detectors composed of the RoI Align layer, two consecutive fully connected layers, namely the RCNN head, and the bounding box classifier and regressor. These detectors have three stages with increasing IoU threshold $\alpha_t$ at each stage $t$, where the following stages take the output of a previous stage as their input. Predicted bounding boxes from the regressor of each stage show necessarily better output IoU than the input IoU, yet for further dynamic improvement, the bounding boxes go through an additional refinement process. This successive balancing process enables the detector to obtain sufficient positive samples at all different IoU levels, which prevents overfitting to the predominant low-quality proposals.
In the base training, we fully train the whole network on the $C_{base}$. At each stage $t$, classifier $\textit{l}_t$ and regressor $\textit{g}_t$ is optimized for IoU threshold $\alpha_t$.
\begin{equation}
\mathcal{L}_{rcnn_t}=\mathcal{L}^{cls}_{rcnn}(l_t(X_i),c_i)+\mathcal{L}^{reg}_{rcnn}(g_t(X_i, p_i),b_i)
\end{equation}
,where $p_i$ and $X_i$ denotes the predicted bounding boxes and the extracted features from the $i$-th object out of N object instances in the input image $x$, respectively. While $T$ denotes the total number of stages, the regression function and the loss for the base training stage can be formulated as 
\begin{equation}
    g(X_i,p_i)=g_T\circ g_{T-1}\circ \cdot\cdot\cdot \circ g_1(X_i,p_i)
\end{equation}
\begin{equation}
\mathcal{L}_{total}=\mathcal{L}_{rpn}^{cls}+\mathcal{L}_{rpn}^{reg}+\sum^T_{t=1}\lambda_t(\mathcal{L}_{rcnn}^{cls}+\mathcal{L}_{rcnn}^{reg})
\end{equation} 

where $\lambda_t$ indicates the loss coefficient of the each stage.
For novel fine-tuning, instead of following the previous TFA fine-tuning strategy, we jointly update the RPN with the classifiers and regressors. 
We provide opportunities for RPN to learn the representations of unseen novel classes, and accordingly enlarge the amount of the RoIs fed to the following classifiers and regressors. To further enrich training samples for the back end of the network, we double the total number of RoIs after the Non-maximum suppression (NMS) is conducted. The loss of the RPN during the novel fine-tuning can be formulated as below:
\begin{equation}
\mathcal{L}_{rpn}=\gamma_{rpn} (\mathcal{L}_{rpn}^{cls}+\mathcal{L}_{rpn}^{reg})
\end{equation} 
Nevertheless, as it is also important to avoid deteriorating the pre-trained RPN, we regulate the rpn loss coefficient $\gamma_{rpn}$ and examine the three cases of $\gamma_{rpn}$:
(1) $\gamma_{rpn} =0$ which corresponds to not updating RPN at all, and the gradients being only dominated by the last layer of RCNN during fine-tuning as in the original TFA. 
(2) $\gamma_{rpn} \in (0,1]$ which corresponds to scaling the gradient of RPN with nonzero loss coefficient.
(3) $\gamma_{rpn} =1$ which is equivalent to updating the RPN without considering the base forgetting. 
We empirically scale the gradients with a nonzero coefficient of 0.5. This scaling is also applied during the recognition of the novel classes. In this manner, the base feature space becomes less corrupted while the scaled gradients assist the construction of the novel feature space.
The whole loss for the second stage is as follows:
\begin{equation}
    \mathcal{L}_{total} =\mathcal{L}_{rpn}+\sum^T_{t=1}\lambda_t(\mathcal{L}_{rcnn}^{cls}+\mathcal{L}_{rcnn}^{reg})
\end{equation}
\section{Experiments}
\subsection{Experimental Setting}
We follow the few-shot training protocol from previous works\cite{wang2020frustratingly,zhang2021hallucination,wu2020multi} and assess our framework on the same data splits of PASCAL VOC\cite{everingham2010pascal} and COCO\cite{lin2014microsoft}. PASCAL VOC uses three different random class splits which of each are composed of 15 base classes and 5 novel classes: novel category split 1,2,3. Different to the data-abundant base classes, novel training set consists of \textit{K} = 1, 2, 3, 5, 10  objects sampled from the combination of VOC07 and VOC12 train/val set. For evaluation, we utilize VOC07 test set and report the default PASCAL Challenge protocol mAP50 and the 11-point interpolated AP of IoU range [0.5,1.0], such that we can evaluate the framework under comprehensive conditions. 
\begin{table*}
\centering
\small
\setlength\extrarowheight{0.3pt}
\setlength{\tabcolsep}{5.8pt}
\caption{Few-shot detection performance (AP50) on PASCAL VOC novel classes of three category splits. Results in \textcolor{red}{red} and \textcolor{blue}{blue} denotes the first and the second best, respectively. Results in \textbf{bold} denotes the higher one compared to the main baseline.}
\label{tab:main-table}
\begin{tabular}{c|ccccc|ccccc|ccccc} 
\toprule
\multirow{2}{*}{Method}                           & \multicolumn{5}{c|}{Novel Category Set 1}                                                                                                                               & \multicolumn{5}{c|}{Novel Category Set 2}                                                                                                                              & \multicolumn{5}{c}{Novel Category Set 3}                                                                                                                             \\
                                                  & shot=1                          & 2                               & 3                               & 5                               & 10                              & shot=1                          & 2                               & 3                               & 5                               & 10                              & shot=1                          & 2                              & 3                              & 5                              & 10                              \\ 
\midrule\midrule
\rowcolor[rgb]{0.937,0.937,0.937} \textbf{ Ours } & 39.2                            & \textbf{\textcolor{blue}{49.2}} & \textbf{\textcolor{red}{56.4}}  & \textbf{57.4}                   & \textbf{\textcolor{red}{61.6}}  & \textbf{28.7}                   & \textbf{31.3}                   & \textbf{36.9}                   & \textbf{37.4}                   & \textbf{44.3}                   & \textbf{37.4}                   & \textbf{\textcolor{red}{44.3}} & \textbf{\textcolor{red}{48.6}} & \textbf{\textcolor{red}{51.8}} & \textbf{\textcolor{red}{56.0}}  \\
TFA w/cos \cite{wang2020frustratingly}                                         & \textbf{39.8}                   & 36.1                            & 44.7                            & \textbf{\textcolor{blue}{55.7}} & 56.0                            & 23.5                            & 26.9                            & 34.1                            & 35.1                            & 39.1                            & 30.8                            & 34.8                           & 42.8                           & 49.5                           & 49.8                            \\ 
\midrule
FRCN+ft-full\cite{wang2019meta}                                      & 15.2                            & 20.3                            & 29.0                            & 40.1                            & 45.5                            & 7.9                             & 15.3                            & 26.2                            & 31.6                            & 39.1                            & 9.8                             & 11.3                           & 19.1                           & 35.0                           & 45.1                            \\
Meta-Det\cite{yan2019meta}                                          & 18.9                            & 20.6                            & 30.2                            & 36.8                            & 49.6                            & 21.8                            & 23.1                            & 37.8                            & 31.7                            & 43.0                            & 20.6                            & 23.9                           & 29.4                           & 43.9                           & 44.1                            \\
Meta R-CNN\cite{wang2019meta}                                         & 19.9                            & 25.5                            & 35.0                            & 45.7                            & 51.5                            & 10.4                            & 19.4                            & 29.6                            & 34.8                            & 45.4                            & 14.3                            & 18.3                           & 27.5                           & 41.2                           & 48.1                            \\
TFA w/fc\cite{wang2020frustratingly}                                          & 36.8                            & 29.1                            & 43.6                            & 55.7                            & 57.0                            & 18.2                            & 29.0                            & 33.4                            & 35.5                            & 39.0                            & 27.7                            & 33.6                           & 42.5                           & 48.7                           & 50.2                            \\
MPSR\cite{wu2020multi}                                              & 41.7                            & 43.1                            & \textbf{\textcolor{blue}{51.4}} & 55.2                            & 61.8                            & 24.4                            & 29.5                            & 39.2                            & 39.9                            & \textbf{\textcolor{blue}{47.8}} & 35.6                            & 40.6                           & 42.3                           & 48.0                           & 49.7                            \\
RetentiveR-CNN\cite{fan2021generalized}                                    & 42.4                            & 45.8                            & 45.9                            & 53.7                            & 56.1                            & 21.7                            & 27.8                            & 35.2                            & 37.0                            & 40.3                            & 30.2                            & 37.6                           & 43.0                           & 49.7                           & 50.1                            \\
FsDetView\cite{xiao2020few}                                         & 24.2                            & 35.3                            & 42.2                            & 49.1                            & 57.4                            & 21.6                            & 24.6                            & 31.9                            & 37.0                            & 45.7                            & 21.2                            & 30.0                           & 37.2                           & 43.8                           & 49.6                            \\
NP-RepMet\cite{yang2020restoring}                                         & 37.8                            & 40.3                            & 41.7                            & 47.3                            & 49.4                            & \textbf{\textcolor{red}{41.6}}  & \textbf{\textcolor{red}{43.0}}  & \textbf{\textcolor{red}{43.4}}  & \textbf{\textcolor{red}{47.4}}  & \textbf{\textcolor{red}{49.1}}  & 33.3                            & 38.0                           & 39.8                           & 41.5                           & 44.8                            \\
CoRPNs+Halluc\cite{zhang2021hallucination}                                     & \textbf{\textcolor{blue}{47.0}} & 44.9                            & 46.5                            & 54.7                            & 54.7                            & 26.3                            & 31.8                            & 37.4                            & 37.4                            & 41.2                            & \textbf{\textcolor{red}{40.4}}  & \textbf{\textcolor{blue}{42.1}}                           & 43.3                           & \textbf{\textcolor{blue}{51.4}} & 49.6                            \\
CME\cite{li2021beyond}                                               & 41.5                            & 47.5                            & 50.4                            & \textbf{\textcolor{red}{58.2}}  & \textbf{\textcolor{blue}{60.9}} & 27.2                            & 30.2                            & \textbf{\textcolor{blue}{41.4}} & \textbf{\textcolor{blue}{42.5}} & 46.8                            & 34.3                            & 39.6                           & \textbf{\textcolor{blue}{45.1}}                           & 48.3                           & \textbf{\textcolor{blue}{51.5}}                            \\
SRR-FSD\cite{zhu2021semantic}                                           & \textbf{\textcolor{red}{47.8}}  & \textbf{\textcolor{red}{50.5}}  & 51.3                            & 55.2                            & 56.8                            & \textbf{\textcolor{blue}{32.5}} & \textbf{\textcolor{blue}{35.3}} & 39.1                            & 40.8                            & 43.8                            & \textbf{\textcolor{blue}{40.1}} & 41.5                           & 44.3                           & 46.9                           & 46.4                            \\
\bottomrule
\end{tabular}
\end{table*}
\begin{table}
\centering
\small
\setlength\extrarowheight{0.3pt}
\setlength{\tabcolsep}{6pt}
\caption{Few-shot detection performance (AP) on PASCAL VOC novel classes of the second base and novel category split. Results in \textbf{bold} denotes the higher one compared to the main baseline. AP is calculated at 0.05 intervals at 11-point interpolated in IoU range from 0.5 to 1.0.}
\label{tab:ap_comparison}
\begin{tabular}{c|ccccc} 
\toprule
\multirow{2}{*}{\begin{tabular}[c]{@{}c@{}}Method\end{tabular}} & \multicolumn{5}{c}{Novel AP}                                                                                                                                                                                                                                                                                                                                                                                                                                                                                           \\
                                                                  & shot=1                                                                                               & 2                                                                                                     & 3                                                                                                    & 5                                                                                                    & 10                                                                                                    \\ 
\midrule\midrule
TFA w/cos\cite{wang2020frustratingly}                                                         & 23.4                                                                                                 & 21.5                                                                                                  & 27.8                                                                                                 & 34.2                                                                                                 & 34.5                                                                                                  \\
\rowcolor[rgb]{0.937,0.937,0.937} Ours                           & \begin{tabular}[c]{@{}>{\cellcolor[rgb]{0.937,0.937,0.937}}c@{}}\textbf{27.8}\\\textbf{(+4.4)}\\\end{tabular} & \begin{tabular}[c]{@{}>{\cellcolor[rgb]{0.937,0.937,0.937}}c@{}}\textbf{33.0}\\\textbf{(+11.5)}\\\end{tabular} & \begin{tabular}[c]{@{}>{\cellcolor[rgb]{0.937,0.937,0.937}}c@{}}\textbf{37.2}\\\textbf{(+9.4)}\\\end{tabular} & \begin{tabular}[c]{@{}>{\cellcolor[rgb]{0.937,0.937,0.937}}c@{}}\textbf{37.5}\\\textbf{(+3.3)}\\\end{tabular} & \begin{tabular}[c]{@{}>{\cellcolor[rgb]{0.937,0.937,0.937}}c@{}}\textbf{41.4}\\\textbf{(+6.9)}\\\end{tabular}  \\
\bottomrule
\end{tabular}
\end{table}
\begin{table}
\centering
\small
\setlength\extrarowheight{0.3pt}
\setlength{\tabcolsep}{5pt}
\caption{The 10-shot cross-domain few-shot detection performance on COCO base set → PASCAL VOC novel set. We follow the evaluation setting from \cite{wang2019meta,wu2020multi}}
\label{tab:cross-domain}
\begin{tabular}{c|cccc|c} 
\toprule
Method & FRCN-ft & MetaDet & MetaRCNN & MPSR& Ours                                               \\ 
\midrule\midrule
mAP    & 31.2    & 33.9    & 37.4     & 42.3 & {\cellcolor[rgb]{0.937,0.937,0.937}}\textbf{48.9}  \\
\bottomrule
\end{tabular}
\end{table}
\begin{table*}
\centering
\small
\setlength\extrarowheight{0.3pt}
\setlength{\tabcolsep}{4.1pt}
\caption{ Effectiveness of the proposed ideas in our approach. All results are conducted on the second split of PASCAL VOC dataset and evaluated based on the AP50 of novel and base classes, respectively and 11-point interpolated AP of novel classes as well.}
\label{tab:ablation}
\begin{tabular}{c|cc|ccccc|ccccc|ccccc} 
\toprule
\multirow{2}{*}{Method} & \multirow{2}{*}{\begin{tabular}[c]{@{}c@{}}Proposal\\ Refinement\end{tabular}} & \multirow{2}{*}{\begin{tabular}[c]{@{}c@{}}Fine-tune \\ RPN\end{tabular}} & \multicolumn{5}{c|}{Novel AP50}                                               & \multicolumn{5}{c|}{Base AP50}                                                & \multicolumn{5}{c}{Novel AP}                                                            \\
                        &                                                                                &                                                                           & 1             & 2             & 3             & 5             & 10            & 1             & 2             & 3             & 5             & 10            & 1                      & 2             & 3             & 5             & 10             \\ 
\midrule\midrule
TFA w/cos\cite{wang2020frustratingly}               &  \xmark                                                                              &   \xmark                                                                        & 23.5          & 26.9          & 34.1          & 35.1          & 39.1          & 79.5          & 77.7          & 78.8          & 78.9          & 78.5          & 11.7                   & 14.0          & 17.9          & 19.9          & 21.1           \\ 
\midrule
\multirow{2}{*}{Ours}   &    \cmark                                                                            &  \xmark                                                                         & 25.2          & 29.8.         & 34.6          & 33.7          & 42.6          & \textbf{81.1} & 80.3          & \textbf{80.6} & \textbf{81.4} & \textbf{81.0} & 15.0                   & 19.3          & 22.7          & 21.7          & 26.5           \\
                        &  \cellcolor[rgb]{0.937,0.937,0.937}\cmark                                                                              &   \cellcolor[rgb]{0.937,0.937,0.937}\cmark                                                                        & \cellcolor[rgb]{0.937,0.937,0.937}\textbf{28.7} & \cellcolor[rgb]{0.937,0.937,0.937}\textbf{31.3} & \cellcolor[rgb]{0.937,0.937,0.937}\textbf{36.9} & \cellcolor[rgb]{0.937,0.937,0.937}\textbf{37.4} & \cellcolor[rgb]{0.937,0.937,0.937}\textbf{44.3} & 
                        \cellcolor[rgb]{0.937,0.937,0.937}80.3          & \cellcolor[rgb]{0.937,0.937,0.937}\textbf{80.4} & 
                        \cellcolor[rgb]{0.937,0.937,0.937}80.5          & 
                        \cellcolor[rgb]{0.937,0.937,0.937}80.6          & 
                        \cellcolor[rgb]{0.937,0.937,0.937}80.6          & 
                        \cellcolor[rgb]{0.937,0.937,0.937}\textbf{\textbf{17.3}} & \cellcolor[rgb]{0.937,0.937,0.937}\textbf{20.1} & \cellcolor[rgb]{0.937,0.937,0.937}\textbf{23.3} & \cellcolor[rgb]{0.937,0.937,0.937}\textbf{24.2} & \cellcolor[rgb]{0.937,0.937,0.937}\textbf{26.9}  \\
\bottomrule
\end{tabular}
\end{table*}
\subsection{Existing Baselines}
As our approach is focusing on resolving the remaining issue of TFA, the proposed method is compared with the main baseline TFA w/cos\cite{wang2020frustratingly} and the other approaches\cite{yan2019meta,wang2019meta,xiao2020few,yang2020restoring,zhu2021semantic,wu2020multi,fan2021generalized,zhang2021hallucination,li2021beyond} to verify the effectiveness of our approach. 
\subsection{Implementation Details}
We use ImageNet\cite{deng2009imagenet} pretrained ResNet-101\cite{he2016deep} with FPN\cite{lin2017feature} as a feature extractor. All models are trained using the SGD optimizer with the batch size of 8, the momentum of 0.9 and the weight decay of 0.0001. The learning rate is set to 0.02 during the base training phase and 0.01 during the few-shot fine-tuning phase. 
IoU threshold $\alpha$ is set to 0.5, 0.6, 0.7 and R-CNN loss coefficient $\lambda$ is set to 1, 0.5, 0.25  at each stage, respectively. The RPN loss coefficient during the fine-tuning stage $\gamma_{rpn}$ is set to 0.5.
\begin{figure}
\begin{center}
\includegraphics[width=1\linewidth]{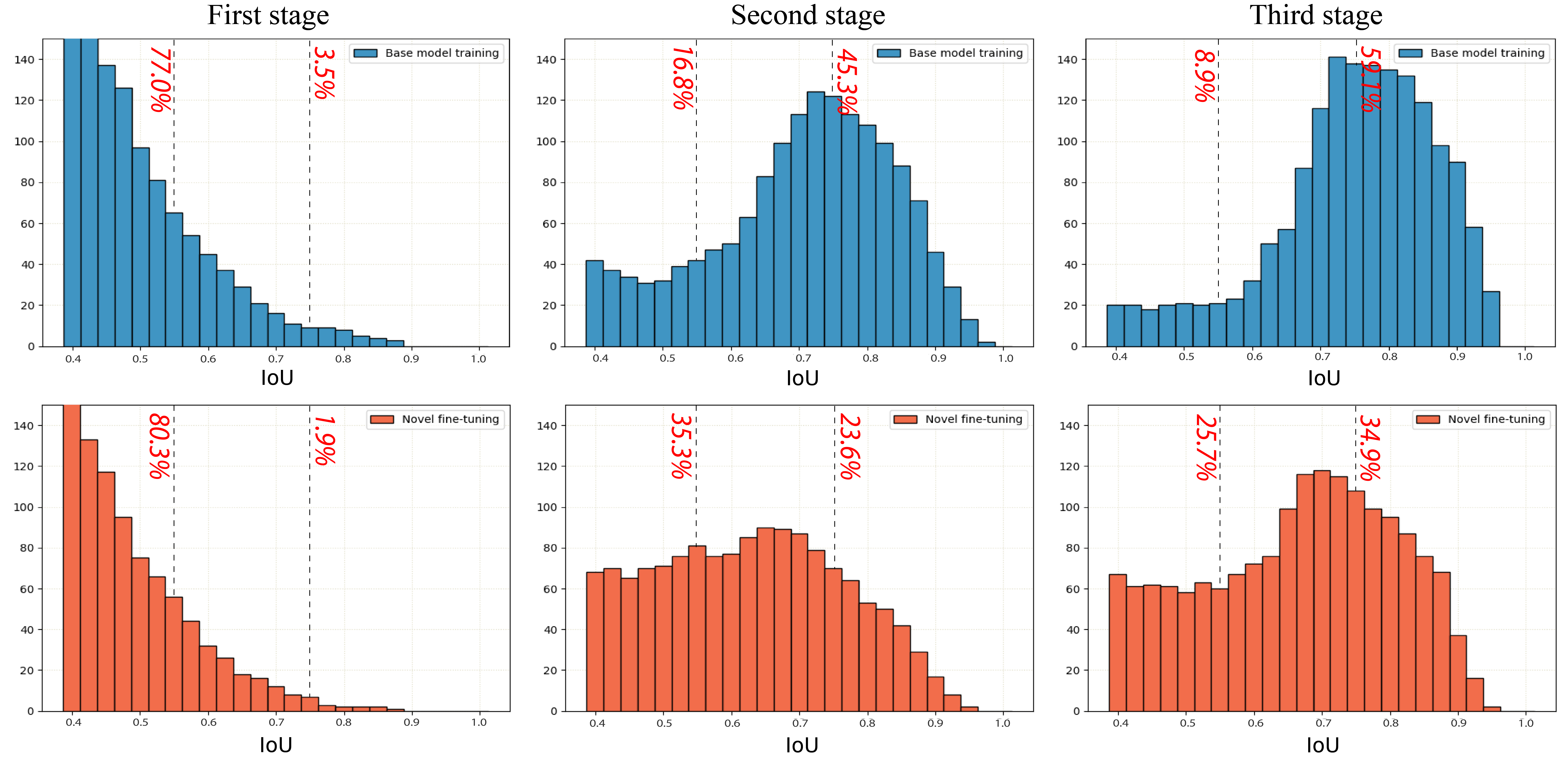}
\end{center}
\caption{Improvement of IoU distribution at each stage. Blue and orange denotes the base training and novel fine-tuning, respectively. IoU distribution of both stages shows similar statistics.}
\end{figure}
\subsection{Benchmarking Results}
We present our main evaluation results of PASCAL VOC on three different data splits in Table \Romannum {1}. The proposed framework significantly outperforms the main baseline, and is mostly superior or comparable to other state-of-the-art baselines. In particular, we demonstrate the effectiveness of our approach by achieving an increase in performance of up to 11.7\% on the main baseline, and 5\% on the current state-of-the-art.
Based on the Table \Romannum {1}, we notice that although the proposed method significantly surpasses the main baseline, the performance gain differs based on the selection of classes and the number of shots. This appears on other baselines as well, which can be explained by the inclusion of low quality samples of specific classes.
For more detailed comparison, we further assess the detector on the 11-point AP from range 0.5 to 1.0 as in \cite{Detectron2018}. We compare the performance with the main baseline as the general AP is only reported in TFA. Table \Romannum{2} shows that our model achieves significant improvements in every case of up to 11.5\%. 
Furthermore, we have two other notable observations: (1) We achieve 79.1\%, 80.1\%, 80.1\% on base classes of three splits respectively before the novel fine-tuning, which are lower than the TFA counterpart 80.8\%, 81.9\%, 82.0\%. Nevertheless, after the novel fine-tuning, as shown in the Table \Romannum{1}, our approach significantly outperforms TFA. It is clear that addressing the imbalanced IoU distribution is a particularly effective solution for alleviating the restricted performance of few-shot object detection. 
(2) IoU distribution of novel RoIs tilts more strongly towards high quality samples after each resampling stage.
Fig.4 shows that among the RoIs of IoU $\geq$ 0.4, high quality RoI samples of IoU $\geq$ 0.75 comprise only 3.5\% and 1.9\% at the first stage, but they increase to 59.1\% and 34.9\% at the final stage in base training and novel fine-tuning, respectively. The optimized IoU threshold is more obvious in the base training, however, note that this is fully due to the great abundance of data. 
For qualitative comparison, we provide visualized results of detected novel objects for 5-shot case of the main baseline and our approach in Fig. 5. Our model produces better detection results on 4 different error cases of mis-localization, mis-classification, missing objects and the complex error case.

\subsection{Results on COCO to VOC}
We conduct the cross-domain few-shot object detection experiment following the previous works\cite{wang2019meta, wu2020multi}. We followed the general experimental setting where $C_{base}$ is the 60 classes in COCO\cite{lin2014microsoft} and $C_{novel}$ is the 20 classes in PASCAL VOC that are not overlapped with $C_{base}$. $S_{novel}$ is composed of 10-shot objects for each of $C_{novel}$. As shown in the Table \Romannum {3}, our approach achieves the highest performance of 48.9\% which is more than 6.5\% AP gain compared to the previous best performance. This leap in performance verifies that our approach has stronger generalization ability when the characteristics of the domains are different. 

\begin{figure}
\begin{center}
\small
\includegraphics[width=1\linewidth]{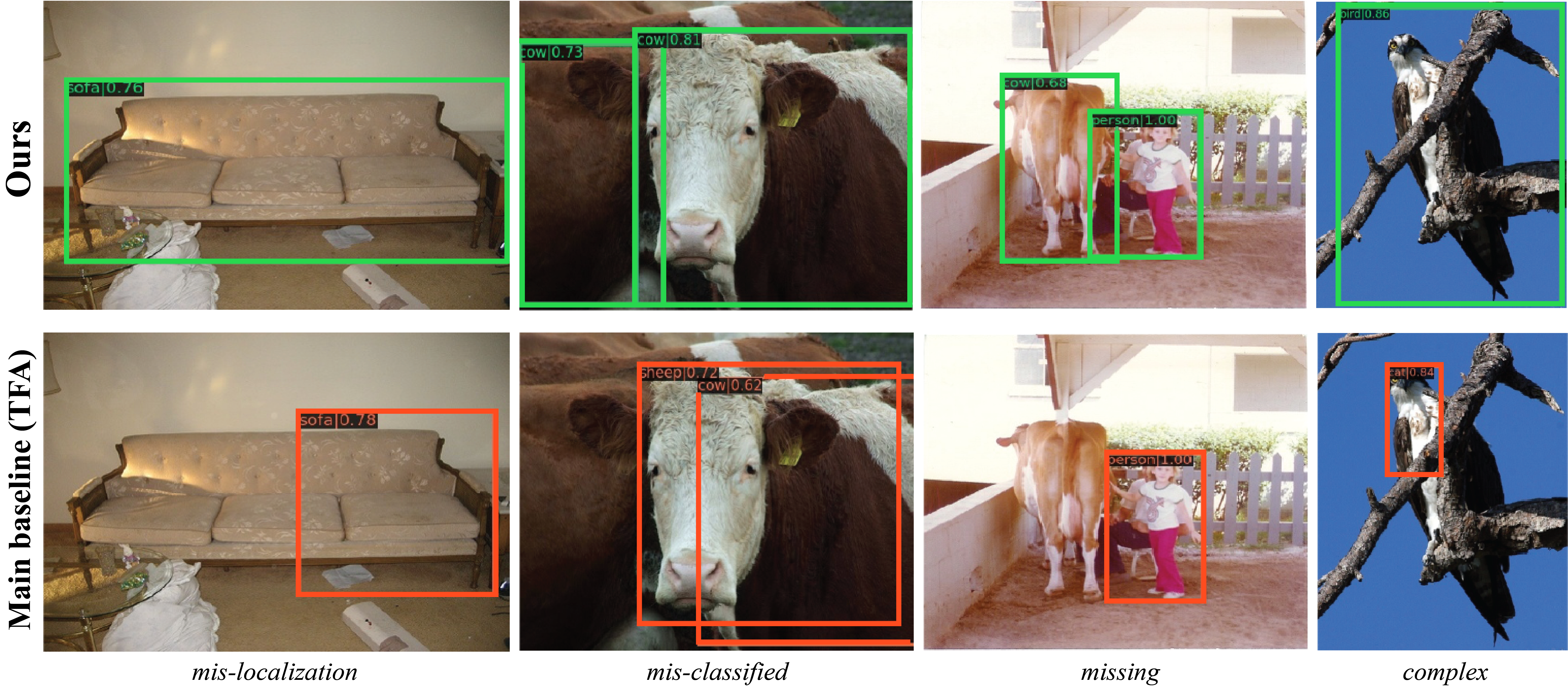}
\end{center}
\caption{Qualitative comparison between our approach and the main baseline on PASCAL VOC 5-shot case on 4 different various error cases of mis-localization, mis-classification, missing objects and complex case of multiple kinds of error.}
\end{figure}

\begin{table}
\centering
\small
\setlength\extrarowheight{0.3pt}
\setlength{\tabcolsep}{6pt}
\caption{ Recall change based on fine-tuning RPN. The assessment is conducted on both novel classes and the balanced set of base and novel classes.}
\label{tab:ablation}
\begin{tabular}{c!{\vrule width \lightrulewidth}l!{\vrule width \lightrulewidth}cc} 
\toprule
\multicolumn{2}{c!{\vrule width \lightrulewidth}}{\multirow{2}{*}{Method / Metric}} & \multicolumn{2}{c}{Recall@100}                    \\
\multicolumn{2}{c!{\vrule width \lightrulewidth}}{}                                 & $C_{novel}$      & $C_{base} \cup C_{novel}$  \\ 
\midrule\midrule
\multirow{2}{*}{Fine-tune~RPN} & \xmark                              & 90.8          & 95.6                                  \\
                               & \cellcolor[rgb]{0.937,0.937,0.937}\cmark                              & \cellcolor[rgb]{0.937,0.937,0.937}\textbf{92.2} & \cellcolor[rgb]{0.937,0.937,0.937}\textbf{95.9}                                  \\
\bottomrule
\end{tabular}
\end{table}

\subsection{Ablation Study}
To verify the effectiveness of our approach, a comprehensive analysis is conducted on the every case of \textit{K} = 1, 2, 3, 5, 10 of the second split of PASCAL VOC.

\subsubsection{Proposal Balance Refinement}
Table \Romannum{4} shows that refining the object candidates can seamlessly improve both AP50 and AP of novel classes compared to the main baseline, upto 5.2\% and 6.1\%, respectively. We observe that improvement of the AP is more significant than in the AP50, as our initial motivation was to improve the impaired RoIs of unseen novel classes. 
\subsubsection{Fine-tuning RPN}
The effectiveness of fine-tuning RPN can be seen in Table \Romannum{4} and Table \Romannum{5}. Table \Romannum{4} shows excluding the RPN update consistently degrades the performance on both AP50 and AP of novel classes but slightly improves the AP50 of base classes. Comparing the largest degree of the improvement of novel classes, 3.7\%, and the degradation of base classes, 0.8\%, it is clear that fine-tuning RPN is a effective way to improve the novel detector while doing less harm to the previously well-established detection capability. The same conclusion can be also drawn from Table \Romannum{5} by comparing the improvement of recall of RPN as we achieve an 1.4\% performance gain on novel unseen classes.

\section{Conclusion}
This paper targets the imbalanced IoU distribution which has been rarely handled, but has a crucial role as a fundamental basis in few-shot object detection. We present a new approach of balancing out the biased IoU distribution via auxiliary refinement. It corrects the proposals of low quality with sequential regression refinement, thus increasing the amount of positive samples of different IoU degrees. 
We further revise the fine-tuning strategy and expose RPN to the novel classes to expand the learning opportunities of the RoI classifier and regressor. Extensive experiments conducted on PASCAL VOC and COCO demonstrate the advantage of our model compared to other state-of-the-art methods. 

\bibliographystyle{IEEEtran.bst}
\bibliography{gu.bib}
\end{document}